\pdfoutput=1
\documentclass{article}

\usepackage{xcolor}
\usepackage[utf8]{inputenc} % allow utf-8 input
\usepackage[T1]{fontenc}    % use 8-bit T1 fonts
\usepackage{url}            % simple URL typesetting
\usepackage{booktabs}       % professional-quality tables
\usepackage{nicefrac}       % compact symbols for 1/2, etc.
\usepackage{microtype}
\usepackage{amsfonts}       % blackboard math symbols
\usepackage{caption}
\usepackage{subcaption}
\usepackage{hyperref}

\usepackage{graphicx}
\usepackage{amsmath}
\PassOptionsToPackage{numbers, compress}{natbib}
\usepackage[final]{neurips_2019}

\title{Graph Generation with Variational Recurrent Neural Network}

\author{%
  David S.~Hippocampus\thanks{Use footnote for providing further information
    about author (webpage, alternative address)---\emph{not} for acknowledging
    funding agencies.} \\
  Department of Computer Science\\
  Cranberry-Lemon University\\
  Pittsburgh, PA 15213 \\
  \texttt{hippo@cs.cranberry-lemon.edu} \\
}

\author{
  Shih-Yang Su\\
  Virginia Tech \& Borealis AI\\
  \texttt{shihyang@vt.edu} \\
  \And
  Hossein Hajimirsadeghi \\
  Borealis AI \\
  \texttt{hossein.hajimirsadeghi@borealisai.com} \\
  \AND
  Greg Mori \\
  Simon Fraser University \& Borealis AI \\
  \texttt{mori@cs.sfu.ca} \\
}

\begin{document}
\definecolor{bluex}{HTML}{007FFF}

\setlength{\abovedisplayskip}{3pt}
\setlength{\belowdisplayskip}{3pt}
\maketitle
\newcommand{\red}[1]{
{\color{red}{#1}}
}
\maketitle
\newcommand{\blue}[1]{
{\color{bluex}{#1}}
}
\begin{abstract}
Generating graph structures is a challenging problem due to the diverse representations and complex dependencies among nodes. In this paper, we introduce Graph Variational Recurrent Neural Network (GraphVRNN), a probabilistic autoregressive model for graph generation. Through modeling the latent variables of graph data, GraphVRNN can capture the joint distributions of graph structures and the underlying node attributes. We conduct experiments on the proposed GraphVRNN in both graph structure learning and attribute generation tasks. The evaluation results show that the variational component allows our network to model complicated distributions, as well as generate plausible structures and node attributes. 
%, which can learn the joint distribution of graph structures and the underlying features.
% GraphVRNN incorporates a variational auto-encoder with learned prior networks to model informative latent representations of the graph and capture stochasticity and diversity. 
\end{abstract}
\vspace{-5.1mm}
\section{Introduction}
Leveraging the structural relationship over data points have been proven to be beneficial for problems in different research domains~\cite{hamilton2017inductive,kipf2016semi,madjiheurem2019representation,norcliffe2018learning,teney2017graph}. Despite the great success of graph neural networks in graph embedding and supervised settings, graph generation remains a challenging problem. To learn and generate graph structures, a generative graph model needs to capture the subtle relationships between node entities. Prior works have used variational Bayes model~\cite{kingma2013auto,sohn2015learning}, normalizing flow~\cite{dinh2016density}, or generative adversarial network~\cite{goodfellow2014generative} to learn the complex underlying distributions of such relationships, and achieve promising results~\cite{jin2018junction,kipf2016variational,li2018learning,liu2019graph,simonovsky2018graphvae}. %Although these models are capable of learning complicated distributions, they are only applicable to small scale graphs due to their limited scalability. 
However, these models are only applicable to small scale graphs due to their limited scalability. To cope with this issue, Graph Recurrent Neural Network (GraphRNN)~\cite{you2018graph} simplified the generation problem as a sequence prediction problem. GraphRNN cuts down the computational cost by mapping graphs into sequences such that the model only has to consider a subset of nodes during edge generation. While achieving successful results in learning graph structures, GraphRNN cannot faithfully capture the distribution of node attributes (Section~\ref{sec:exp}).
%By making use of strong deep autoregressive model, GraphRNN can learn the patterns among graph data and generate plausible graph with less computational cost.
%However, the model is designed to learn only the structural relationship among data, and cannot faithfully generate the attributes for the graph as we shown in Section~\ref{sec:exp}. \\

In this paper, we propose a novel method for graph representation and generation to address the shortcomings explained above. Our contributions are summarized as follows: (1) A new formulation of an autoregressive variational autoencoder (VAE)~\cite{chung2015recurrent,fabius2014variational,kingma2013auto,mehrasa2019variational} for graph generation which yields a high-capacity model to capture latent factors over graphs with diverse modalities and complicated structural dependencies, while being as scalable as GraphRNN. (2) An algorithm to learn the joint probability distribution of the graph structure as well as the node attributes. (3) Experimental studies with successful results, competitive to the state-of-the-art models for graph generation.

\section{The Proposed Method}
\label{sec:method}
In this section, we first introduce the problem and then describe our proposed GraphVRNN model. Please refer to Appendix~\ref{sec:background} for the background on VAE and GraphRNN.
\subsection{Problem Definition}
Let $G=(A, X)$ be an undirected graph with $n$ nodes, where $A\in \mathbb{B}^{n\times n}$ is the adjacency matrix and $X\in \mathbb{R}^{n\times k}$ is the attribute matrix. Our goal is to model the joint probability distribution of the graph structure and the node attributes, i.e., $p(G) = p(A, X)$.

\begin{figure}[t]
\centering
\includegraphics[width=1.0\linewidth]{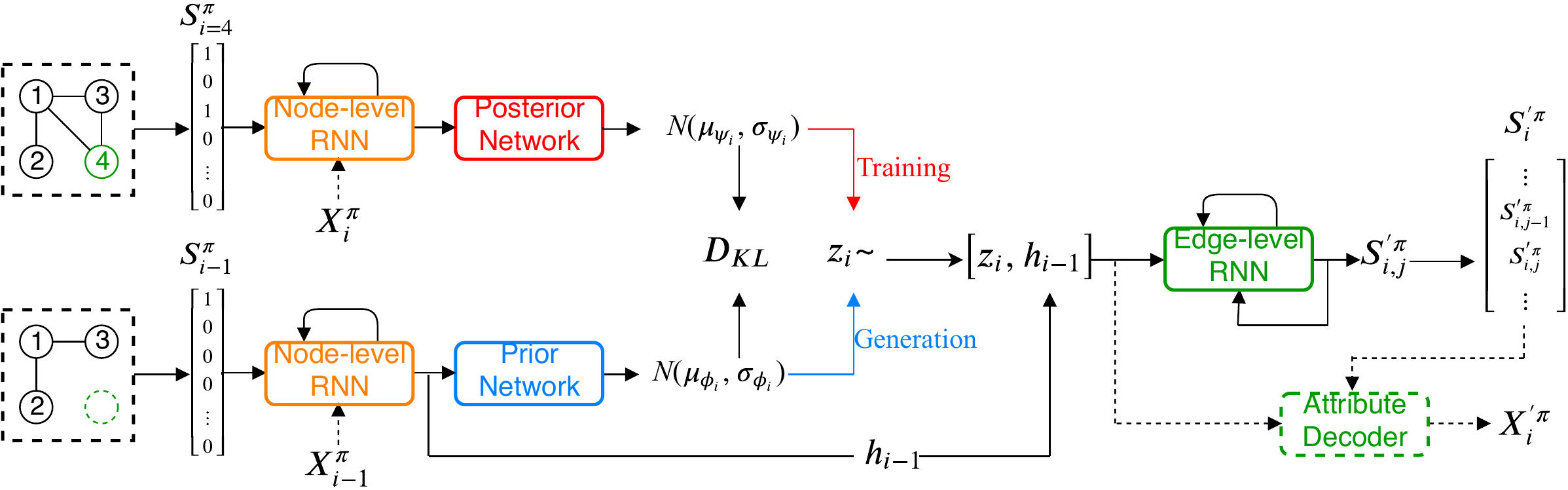}
%\vspace{-1.5em}
\caption{
\textbf{Overview of GraphVRNN.} During \red{training}, the model learns to reconstruct $S^\pi_i$ and $X^\pi_i$ at each step of sequence using the proposal distribution $q_\psi(z_i\vert S^{\pi}_{\leq i}, X^{\pi}_{\leq i})$, while staying close to the learnable prior distribution $p_\phi(z_i\vert S^{\pi}_{<i}, X^{\pi}_{<i})$. During \blue{generation}, the model takes the previously generated $S^\pi_{i-1}$ and $X^\pi_{i-1}$, sample $z_{i}$ from the learned prior and feed $z_i$ and the node-level RNN output $h_{i-1}$ to the decoder to output $S^{\pi}_{i}$ and then $X^{\pi}_{i}$. Black dash arrows indicate the path that is only required for attribute generation. Note that the two node-level RNNs are sharing the same weights.
}
\label{fig:overview}
\end{figure}
\subsection{Graph Recurrent Neural Network with Variational Bayes}
In this section, we propose a VAE formulation to learn a model for $p(G)$, which can represent both graph structure and node attributes. Following the GraphRNN~\cite{you2018graph} approach to graph generation, we map a graph to sequences under Breadth-first Search (BFS) policies:
\begin{equation}
    (S^\pi, X^\pi) = \left(\left[S_1^\pi,S_2^\pi,\cdots,S_n^\pi\right], \left[X_1^\pi,X_2^\pi,\cdots,X_n^\pi\right]\right) = BFS(G, \pi),\label{eq:s_seq}
\end{equation}
where $S^\pi$ is a sequence under a possible node permutation $\pi$ produced by BFS, and $S_i^\pi=\left[S^\pi_{i, 1},S^\pi_{i, 2},\cdots,S^\pi_{i, m}\right]=\left[A^\pi_{  i-1, i},A^\pi_{i-2, i},\cdots,A^\pi_{i-m, i}\right]_{m<n}$ is the adjacency vector (or edge connections) between node $i$ and its $m$ previously generated nodes. Likewise, $X^\pi$ is the sequence of node attribute vectors under the same permutation $\pi$. This sequential setting not only improves scalability but also alleviates the problem with node matching and non-unique representations during training~\cite{you2018graph}. Given this, the objective is reformulated to model $p(S^\pi, X^\pi)$:
\begin{align}
    %&p(G)=\sum_{\pi\in\Pi} p(S^\pi,X^\pi)\mathbf{1}\left[f_G(S^\pi, X^\pi)=G\right],\label{eq:pG2} \nonumber \\
    &p(G)=\sum_{\pi\in\Pi} p(S^\pi,X^\pi),\label{eq:pG2} \nonumber \\
    &p(S^\pi, X^\pi)= p(S_{1:n}^\pi, X_{1:n}^\pi) = \prod_i{p(S_i^\pi, X_i^\pi | S_{<i}^\pi, X_{<i}^\pi)}, 
\end{align}
where $\Pi$ is the set of all possible permutations which gives $G$. Using conditional VAE (CVAE)~\cite{sohn2015learning}  formulation and defining the proposal distributions $q_\psi\left(z_i\vert S^\pi_{\leq i}, X^\pi_{\leq i}\right)$ and prior distributions $p_\phi\left(z_{i}\vert S^\pi_{<i}, X^\pi_{<i}\right)$ for each step in the sequence, the lower bound for the likelihood of $G$ under policy $\pi$, i.e., $\log p(S^\pi, X^\pi)$ is derived as
\begin{align}
    \mathcal{L}_{\theta, \phi, \psi}(S^\pi, X^\pi)&= \sum_i \mathbb{E}_{z_{i}\sim q_\psi(\cdot)}\left[\log p_\theta\left(S^\pi_i, X^\pi_i\vert S^\pi_{<i}, X^\pi_{<i}, z_{\leq i}\right)\right] \nonumber\\
    & -\beta D_{KL}\left(q_\psi\left(z_i\vert S^\pi_{\leq i}, X^\pi_{\leq i}\right)|| p_\phi\left(z_{i}\vert S^\pi_{<i}, X^\pi_{<i}\right)\right) \nonumber\\
    &= \sum_i \mathbb{E}_{z_{i}\sim q_\psi(\cdot)}\left[\log p_{\theta_1}\left(S^\pi_i\vert S^\pi_{<i}, X^\pi_{<i}, z_{\leq i}\right) + \log p_{\theta_2}\left(X^\pi_i\vert S^\pi_{\leq i}, X^\pi_{<i}, z_{\leq i}\right)\right] \nonumber\\
    & -\beta D_{KL}\left(q_\psi\left(z_i\vert S^\pi_{\leq i}, X^\pi_{\leq i}\right)|| p_\phi\left(z_{i}\vert S^\pi_{<i}, X^\pi_{<i}\right)\right),\label{eq:obj}
\end{align}
where $D_{KL}$ is the Kullback–Leibler (KL) divergence and $\beta$ is a hyperparameter for tuning the KL penalty. An overview of the proposed Graph Variational Recurrent Neural Network (GraphVRNN) is illustrated in Figure~\ref{fig:overview}. In this network, a node-level RNN is used to keep the history of $X_i$ and $S_i$. The output is fed to a multi-layer perceptron (MLP) modeling the proposed posterior $q_\psi(z_i\vert S^\pi_{\leq i}, X^\pi_{\leq i})$. At the bottom, the prior $p_{\phi}(z_{i}\vert S^\pi_{<i}, X^\pi_{<i})$ is modeled similarly but the input is one step behind. The proposal and prior distributions can be sampled to produce latent codes $z_i$. Next, $z_i$ and the historical information $h_{i-1}$ (encoded by the node-level RNN) are concatenated and go through the decoding networks $p_{\theta_1}$ and $p_{\theta_2}$ to generate $S'^{\pi}_i$ and then $X'^{\pi}_i$. We use an edge-level RNN as $p_{\theta_1}$ to decode $S^\pi_{i,j}$ step by step. As for the attribute decoder $p_{\theta_2}$, we use an MLP. The model is trained by optimizing Eq.~\ref{eq:obj} using re-parameterization trick~\cite{kingma2013auto}. For implementation details, please refer to Appendix~\ref{sec:details}.

%Decoding $S^\pi_i$ is done by an edge-level RNN that generate $S^\pi_{i,j}$ step by step, while $X^\pi_i$ is decoded by a attribute decoder consists of a 3-layer MLP. We train the model by optimizing Eq.~\ref{eq:obj},  using re-parameterization trick~\cite{kingma2013auto} and approximating the expectation with a single sample.%Decoding $S^{\pi}_i$ can be performed in one shot using an MLP with a sigmoid activation, or through an edge-level RNN to generate $S^{\pi}_{i, j}$ step by step. We train the model by optimizing Eq.~\ref{eq:obj},  using re-parameterization trick~\cite{kingma2013auto} and approximating the expectation with a single sample.
\begin{table}[t]
\footnotesize
\centering
\setlength{\tabcolsep}{3.7pt}
\caption{\textbf{GraphVRNN improves the results on modeling graph structures.} The learned prior helps learning large graphs or graphs with convoluted edge connections. $\vert E\vert$: number of edges.}
\begin{tabular}{lccccccccc}
(max$\vert V\vert$,max$\vert E\vert$)& \multicolumn{3}{c}{Ego-small (18, 69)} & \multicolumn{3}{c}{Com-small (20, 70)} & \multicolumn{3}{c}{Com-mix (40, 198)} \\%& \multicolumn{3}{c}{Com (60, 268)} & \multicolumn{3}{c}{Com-mix (40, 198)} \\
\toprule
& D. & C. & O. & D. & C. & O. & D. & C. & O. \\%& D. & C. & O. \\
\toprule
GraphVAE~\cite{simonovsky2018graphvae} & 0.130 & 0.170 & 0.050 &  0.350 & 0.980 & 0.540 & - & - & - \\%  & - & - & - \\
DeepGMG~\cite{li2018learning} &  0.040 & 0.100 & 0.020 & 0.220 & 0.950 & 0.400 & - & - & - \\%& - & - & -\\
GraphRNN & 0.076 & 0.298 & 0.008 & 0.020 & 0.061 & 0.009 & 0.045 & 0.045 & 0.046 \\%&  0.010 & 0.020 & 0.027  & 0.045 & 0.045 & 0.046\\
\midrule
GraphVRNN-nlp & 0.026 & 0.144 & 0.008 & 0.009 & 0.040 & 0.005 & 0.043 & 0.051 & 0.047 \\ %& 0.010 & 0.026 & 0.029 & 0.043 & 0.051 & 0.047 \\
GraphVRNN & 0.052 & 0.184 & 0.010 & 0.015 & 0.057 & 0.005 & 0.038 & 0.047 & 0.039 \\%& 0.012 & 0.025 & 0.028 & 0.038 & 0.047 & 0.039\\
%GraphVRNN-1.0& 0.060 & 0.175 & 0.008 & 0.015 & 0.058 & 0.005& 0.036 & 0.046 & 0.045 \\%& 0.011 & 0.019 & 0.027  & 0.036 & 0.046 & 0.045\\
\bottomrule
\end{tabular}
\label{tlb:small}
\vspace{-1.9em}
\end{table}

% Adopting the sequence prediction setting similar to GraphRNN, the goal of GraphVRNN is to learn a good approximation of 
\vspace{-0.5mm}
\section{Experiments}
\label{sec:exp}
\vspace{-1mm}
We conduct experiments to evaluate GraphVRNN's ability to (1) model graph structures with complex latent distribution, (2) generate both plausible graph structures and attributes. To examine the effect of the prior network, we include a variant of GraphVRNN with simple $\mathcal{N}(\mathbf{0}, \mathbf{I})$ prior (GraphVRNN-nlp). We adopt the maximum mean discrepancy (MMD) measurement proposed in~\cite{you2018graph} as the evaluation metric. For each dataset, we train all models for 5 runs, and report the averaged MMD of degree (D.), orbits (O.) and clustering coefficient (C.) between the generated examples and the testing set. A lower number implies a better result. For all the datasets considered in our experiments, we use $80\%$ of the graphs as training set, and the remaining $20\%$ as testing set.
\vspace{-1mm}
\subsection{Modeling Graph Structure}
\label{sec:structure}
\vspace{-1mm}
To evaluate the performance of GraphVRNN on learning graph structures, we first consider two small-scale datasets: (1) \textbf{Ego-small}: a 1-hop citation dataset with 200 graphs with $4\leq\vert V\vert\leq 18$ nodes for each graph, and (2) \textbf{Com-small}: 500 2-community graphs with $12\leq \vert V\vert\leq 20$ per graph ($\vert V\vert/2$ nodes in each community). The probability that a pair of nodes is connected is $0.7$ if the nodes are in the same community, and $0.05$ otherwise. 

We present the evaluation results in Table~\ref{tlb:small}. GraphVRNN shows superior results than GraphRNN in all three MMD metrics. Interestingly, GraphVRNN-nlp exhibits better performance than the model with learned prior. This could be due to that the small-scale dataset does not have complex distribution that requires a strong prior for inference. Overall, the results indicate that the variational component in the proposed models help capture both the local and global structures of a small graph dataset.

To further test the efficacy of each model in handling larger graphs, we consider the \textbf{Com-mix} dataset. Similar to Com-small, Com-mix dataset consists of 500 2-community graphs but with $24\leq\vert V\vert\leq40$. In addition, there are two types of graphs in this dataset: one with equal intra-connection probability $0.3$ for both communities, and the other with $0.4$ and $0.6$ for the two communities in the graphs (details in Appendix~\ref{sec:qualitative}). These settings challenge the scalability of the models, and test the models' capability of learning the complex distribution of graph structures and types.  

We show the results in Table~\ref{tlb:small}. One can see that GraphVRNN shows superior performance on modeling the complicated intra-connection probability in Com-mix, which is reflected on the lower degree MMD. In addition, as the training/testing graphs get larger, GraphVRNN starts showing advantages over GraphVRNN-nlp. The results suggest that the learned prior network is helpful for modeling more intricate structures. Finally, we observed that GraphVRNN is insensitive to $\beta$ coeffient in the graph structure learning tasks. 

\vspace{-1mm}
\subsection{Generating Attributes and Structures}
\vspace{-1mm}
\begin{figure}[t]
\begin{minipage}[t]{1.0\linewidth}%
\centering
   \begin{subfigure}[b]{0.5\textwidth}%
        \centering
        \includegraphics[width=1.0\textwidth]{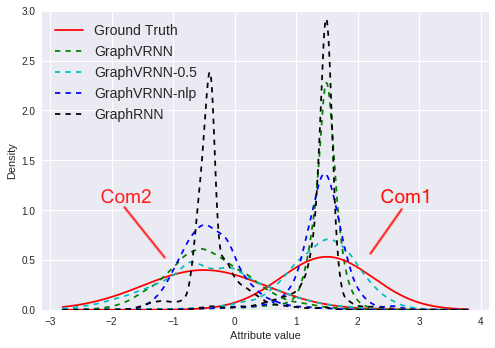}
        %\caption{Community 1}
    \end{subfigure}%
    \begin{subfigure}[b]{0.5\textwidth}%
        \centering
        \includegraphics[width=1.0\textwidth]{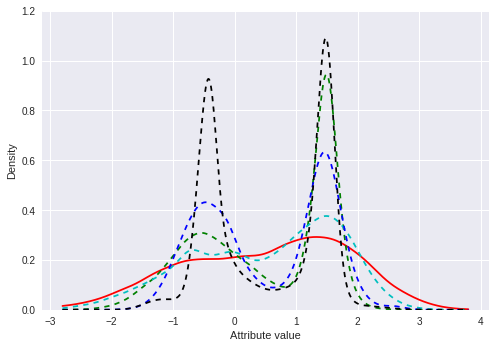}%
        %\caption{Community 2}
    \end{subfigure}%
%\vspace{-0.5em}
\caption{\textbf{Density plot of generated attributes.} We illustrate the true and generated attributes for each community separately (left), and the distribution of all attributes in the whole graph (right).}%
\label{fig:dplot}
\end{minipage}%
\end{figure}
\begin{table}[t]
%\vspace{-1.5em}
\footnotesize
\centering
\caption{\textbf{GraphVRNN performs better on approximating the actual distribution of attributes.} The number after GraphVRNN indicates the $\beta$ coefficient we use for training the model.}
\begin{tabular}{lccccccc}
(max$\vert V\vert$,max$\vert E\vert$)& \multicolumn{3}{c}{Com-attr (60,268)} & & \multicolumn{3}{c}{Earth Mover's Dist.}  \\
\toprule
& D. & C.& O. & & Com1 & Com2 & All \\
\cmidrule[1pt]{1-4}\cmidrule[1pt]{6-8}
GraphRNN & 0.012 & 0.015 & 0.019 & & 0.453 & 0.640 & 0.460 \\
\cmidrule{1-4}\cmidrule{6-8}
GraphVRNN-nlp & 0.017 & 0.015 & 0.018 & & 0.302 & 0.400 & 0.290 \\
GraphVRNN-0.5 & 0.016 & 0.021 & 0.024 & & 0.178 & 0.095 & 0.113 \\
GraphVRNN-1.0 & 0.014 & 0.015 & 0.022 & & 0.382 & 0.265 & 0.267 \\
%GraphVRNN-nlp & 0.010 & 0.026 & 0.029 & 0.230 & 0.368  \\
%GraphVRNN-nlp-0.5 & 0.020& 0.104 & 0.006 & 0.158 & 0.107  \\
%GraphVRNN-0.5 & 0.085 & 0.145 & 0.014 & 0.153 & 0.118 \\
%GraphVRNN-1.0 & 0.076 & 0.177 & 0.019  & 0.347 & 0.216 \\
\bottomrule
\end{tabular}
\vspace{-2em}
\label{tlb:attr}

\end{table}
We have shown that GraphVRNN has comparable or even better ability to model the graph pattern than its non-variational counterpart. Finally, we test GraphVRNN's ability to capture the latent spaces for generating graph structures and attributes simultaneously. Here, we consider \textbf{Com-attr} dataset, a variant of Com-small dataset with even larger graph ($30\leq\vert V\vert\leq 60$ nodes per graph). Moreover, nodes in each of the 2 community draw 1-dimensional attributes from $\mathcal{N}{(1.5, 0.75)}$ and $\mathcal{N}(-0.5, 1.0)$ respectively. Therefore, the models have to construct both the edge connections, as well as generate the corresponding attributes for each community. 

We show the density plot of the generated attributes in each community in Figure~\ref{fig:dplot}, and denote the two communities as Com1 and Com2. One can see that GraphRNN simply generates all the attributes around the modes since the model is trained to maximize the likelihood objective with deterministic recurrent networks, lacking the ability to capture the uncertainty faithfully. On the other hand, GraphVRNN approximates the distribution of Com1 well, while giving reasonable approximates of Com2 (Figure~\ref{fig:dplot}(left)). We further report the Earth Mover's distance (the lower the better) between the generated attribute distribution and the true distribution for (1) each community and (2) the whole graph (denoted as All), along with the MMD metric in Table~\ref{tlb:attr}. It can be seen that GraphRNN yields good results on learning the graph structures, but approximates the attribute distributions poorly. On the other hand, GraphVRNN shows superior results in the attribute modeling task. Comparing GraphVRNN to GraphVRNN-nlp, the results imply that overall the learned prior network helps capture the attribute distribution better. Additionally, $\beta$ has a crucial impact on attribute generation task. Relaxing $\beta$ to 0.5 allows the prior to be more flexible, improving the approximation on the attribute distribution of the whole graph (Figure~\ref{fig:dplot}(right)). To conclude, we show that, equipped with the learned prior network, GraphVRNN is capable of learning graph structures and attribute distributions simultaneously even for large-scale graphs. We include all qualitative results in Appendix~\ref{sec:qualitative}.%Additionally, the results show that for GraphVRNN, $\beta$ has crucial impact on modeling attribute distributions, while being relatively insignificant for structure learning task. Finally, GraphVRNN-nlp does pretty well on both structure learning and attribute modeling. This correspond to our observation in Section~\ref{sec:structure}: a strong prior is not required when the graph size is small. Overall, we show that GraphVRNN and its variants work better for both graph structure and attribute generations than the original GraphRNN, and the prior network could be beneficial for generating larger graphs.
\section{Conclusion}
We proposed GraphVRNN, a variational recurrent network for graph generation. Through modeling the probabilistic latent factors of the graph data, we show that GraphVRNN is capable of modeling the intricate and convoluted distribution of synthetic graph datasets. One future direction of this work is to experiment GraphVRNN in real-world applications (e.g. molecules generation, knowledge graph construction). In contrast to the artificially constructed dataset, the real-world data challenges the model to learn complex or even irregular patterns among both graph structure and attributes, which may necessitate the need of adopting strong priors~\cite{xu2019necessity}. We leave this direction for the future work. %The complicated nature of real-world data may necessitate the need of adopting stronger priors~\cite{xu2019necessity}, and we leave this direction for the future work.

{%\small
%\bibliographystyle{plainnat}
%\bibliography{ref}

}
\clearpage
\appendix
\section{Background}
\label{sec:background}
\subsection{GraphRNN}
\label{sec:graphrnn}
In GraphRNN, they focus on learning the distribution $p(G)$, the structural pattern of graphs without the attribute matrix $X$. They proposed to reformulate the learning task as a sequence prediction problem. First, they map $G$ into a sequence with a breadth-first search (BFS):
\begin{equation}
    S^\pi = BFS(G) = \left[S_1^\pi,S_2^\pi,\cdots,S_n^\pi\right],%\label{eq:s_seq}
\end{equation}
where $S^\pi$ is a sequence under a possible node permutation $\pi$ generated by the BFS policy, and $S_i^\pi=\left[A^\pi_{  i-1, i},A^\pi_{i-2, i},\cdots,A^\pi_{i-m, i}\right]_{m<n}$ is the adjacency vector between node $v_i^\pi$ and its $m$ previously generated nodes. The assumption here is that BFS would map the neighboring nodes of $v_i^\pi$ within $m$ steps, and therefore omits the need of checking all the other $n-1$ nodes. The learning target then becomes $p(S^\pi)$:
\begin{align}
    p(G)&=\sum_{\pi\in\Pi} p(S^\pi)\mathbf{1}\left[f(S^\pi)=G\right],\label{eq:pG} \\
    p(S^\pi)&=\prod_{i=1}^{n} \underbrace{p(S^\pi_i|S^\pi_{<i})}_{\text{node-level RNN}}=\prod_{i=1}^{n}\prod_{j=1}^{m}\underbrace{p(S_{i, j}^\pi|S^\pi_{<i},S^\pi_{i,<j})}_{\text{edge-level RNN}}.\label{eq:pS}
\end{align}
Equation~\ref{eq:pS} is factroized and modeled by a node-level and an edge-level RNN that keeps track of the information on previously generated adjacency vectors $S^\pi_{<i}$ and edges $S^\pi_{i, <j}$ respectively for node $\pi(v_i)$, and $f$ is a function that maps sequence $S^\pi$ back to a graph. The two RNNs predict the adjacency vector $S_i^\pi$ sequentially, and put together the predicted $\left[S^\pi_1,S^\pi_2,\cdots,S^\pi_n\right]$ to generate the adjacency matrix as the final output.

\subsection{Variational Autoencoder}
\label{sec:vae}
Variational Autoencoder (VAE) \cite{kingma2013auto} is a probabilistic generative model with a latent variable $\mathbf{z}$ which can be used to approximate likelihood of data $p(\mathbf{x})$. The generation is performed by sampling from a latent space using the prior distribution $p(\mathbf{z})$. The sampled code $\mathbf{z}$ is next decoded through a conditional distribution $p_{\theta}(\mathbf{x}|\mathbf{z})$ to reconstruct $\mathbf{x}$. Given this, the likelihood of data can be obtained as $p(\mathbf{x}) = \mathbb{E}_{p(\mathbf{z})}p_{\theta}(\mathbf{x}|\mathbf{z})$. However, this is interactable to compute. Thus, $p(\mathbf{x})$ is approximated by a variational lower bound as bellow:
\begin{align}
\label{eq:vae-elbo}
\log p_{\theta}(\mathbf{x}) \geq \mathcal{L}_{\theta, \psi}(\mathbf{x})  = 
  \mathbb{E}_{q_{\psi}(\mathbf{z}|\mathbf{x})}{[\log p_{\theta}(\mathbf{x}|\mathbf{z})} ] -  
{D}_\text{KL}{[q_{\psi}(\mathbf{z}|\mathbf{x}) ||
p(\mathbf{z})]} 
\end{align}
This formulation gives an autoencoder structure composed of an inference network and a generation network to encode and decode data. In the inference network, a proposal posterior $q_{\psi}(\mathbf{z}|\mathbf{x})$ is learned instead of the intractable true posterior $p(\mathbf{z}|\mathbf{x})$ to sample $z$. This sample is next decoded through the generation module $p_{\theta}(\mathbf{x}|\mathbf{z})$ to reconstruct $\mathbf{x}$. $D_\text{KL}$ denotes the Kullback-Leibler (KL) divergence between the prior and the proposal distribution. This can be also interpreted as a regularizer for the reconstruction loss.

In vanilla VAE, the proposal posterior is modeled as diagonal normal distribution and the prior is the standard normal distribution $\mathcal{N}(\mathbf{0}, \mathbf{I})$. This gives an analytical solution for the KL divergence. To train this VAE, $q_{\psi}$ and $p_{\theta}$ are modeled as neural networks, and the parameters $\psi$ and $\theta$ are learned by optimizing~\ref{eq:vae-elbo} using the re-parameterization trick~\cite{kingma2013auto}

Conditional VAE (CVAE)~\cite{sohn2015learning} is an extension to VAE which approximates the conditional distribution $p(\mathbf{x}|\mathbf{y})$.
The variational lower bound for CVAE is defined as:
\begin{flalign}
\log p_{\theta}(\mathbf{x}|\mathbf{y}) \geq \mathcal{L}_{\theta, \phi, \psi}(\mathbf{x}, \mathbf{y})  =   
  \mathbb{E}_{q_{\psi}(\mathbf{z}|\mathbf{x}, \mathbf{y})}{[\log p_{\theta}(\mathbf{x}|\mathbf{z}, \mathbf{y})]}  -  
{D}_\text{KL}{[q_{\psi}(\mathbf{z}|\mathbf{x}, \mathbf{y}) ||
p_{\phi}(\mathbf{z}|\mathbf{y})]} 
\end{flalign}

\section{Model Details}
\label{sec:details}

\subsection{Network Modules}
\label{sec:modules}
% We use binary cross-entropy and mean squared error as the reconstruction losses for $S^{\pi}_i$ and $X^{\pi}_i$ respectively.
For nodel-level RNN and edge-level RNN, we uses a 4-layer GRU unit with a hidden size of 128 and 16 respectively as in~\cite{you2018graph}. The prior and posterior networks both consist of a 3-layer MLP with ReLU non-linearity for hidden layers, and mapped the output of node-level RNN into a 64-dim latent vector $z_i$. Similarly, the attribute decoder adopts a 3-layer MLP network, and take as input the $z_{i-1}$, the hidden state of node-level RNN $h_{i-1}$ and the predicted adjacency vector $S^{\prime\pi}_i$ to generate the note attribute $X_i^{\prime\pi}$. We use binary cross-entropy and mean squared error as the reconstruction losses for $\log p_{\theta_1}(S^{\pi}_i\vert S^\pi_{<i},X^\pi_{<i},z_{\leq i})$ and $\log p_{\theta_2}(X^{\pi}_i\vert S^\pi_{<i},X^\pi_{<i},z_{\leq i})$ respectively. For small size graph, we follow the practice in~\cite{you2018graph} to half the hidden size of the node-level RNN.

For GraphRNN, the input of attribute decoder is simply $h_{i-1}$ and $S^{\prime\pi}_i$.

\subsection{Experiment Settings}
\label{sec:settings}
We use the official GraphRNN Github\footnote{GraphRNN: \url{https://github.com/JiaxuanYou/graph-generation}} repository provided by the authors for training the baseline models, generating the synthetic dataset set and evaluating the MMD metric. We follow the same hyperparameter settings suggested in~\cite{you2018graph}: in all experiments, we use Adam optimizer~\cite{kingma2014adam} with batch size 32 (32 graphs per batch). The learning rate is set to 0.001, and decay it by 0.3 at training steps 12800 and 32000. In all experiments, we only consider $\beta=0.5$ or $1.0$, without doing any hyperparameter search.

\section{Qualitative Result}
\label{sec:qualitative}
\begin{figure}[t]
\begin{minipage}[t]{1.0\linewidth}%
\centering
   \begin{subfigure}[b]{0.5\textwidth}%
        \centering
        \includegraphics[width=1.0\textwidth]{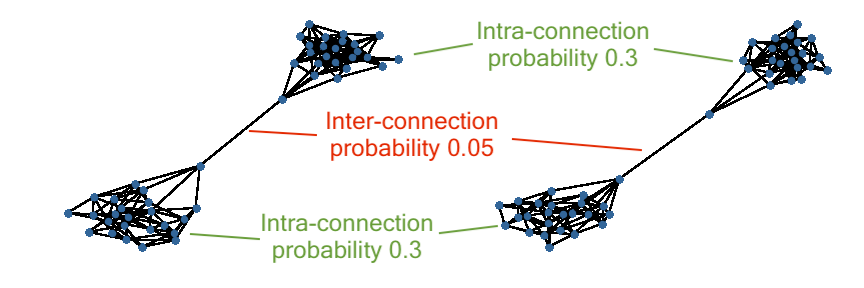}
        \caption{Sample graph structures of Com-attr dataset.}
        \label{fig:comattr}
    \end{subfigure}%
    \begin{subfigure}[b]{0.5\textwidth}%
        \centering
        \includegraphics[width=1.0\textwidth]{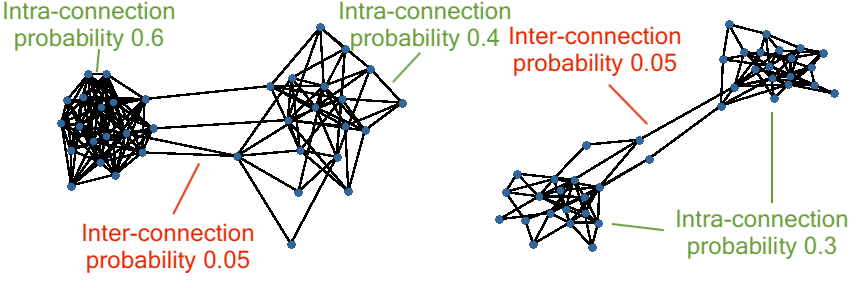}%
        \caption{Sample graph structures of Com-mix dataset.}
        \label{fig:commix}
    \end{subfigure}%
\caption{\textbf{Illustration of graph samples from Com-attr (a) and Com-mix datset (b).} For Com-attr (a) and Com-small, the intra-connection probability for both communities are the same for every graph in the dataset. On the other hands, two types of graphs exist in Com-mix (b): one with different intra-connection probability for the two communities ((b) left) and one with the same intra-connection probability ((b) right).}%
\label{fig:dataset}
\end{minipage}%
\end{figure}
\newcommand{\mfigure}[2]{
{\includegraphics[#1]{#2}}
}
\begin{figure}[t]
\centering
\begin{tabular}{ccc}
        \multicolumn{3}{c}{Ego-small} \\
         \mfigure{width=0.25\linewidth}{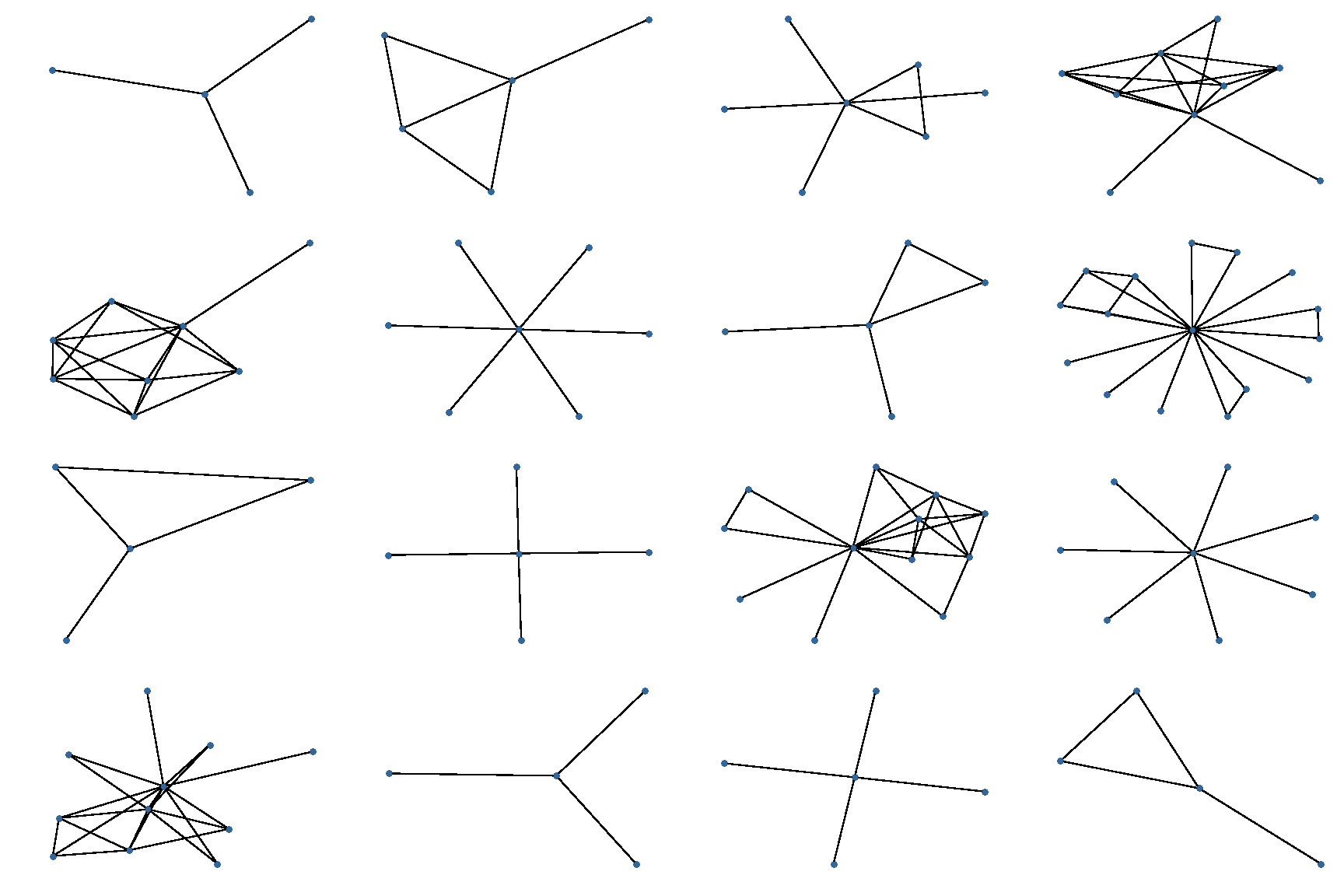} & \mfigure{width=0.25\linewidth}{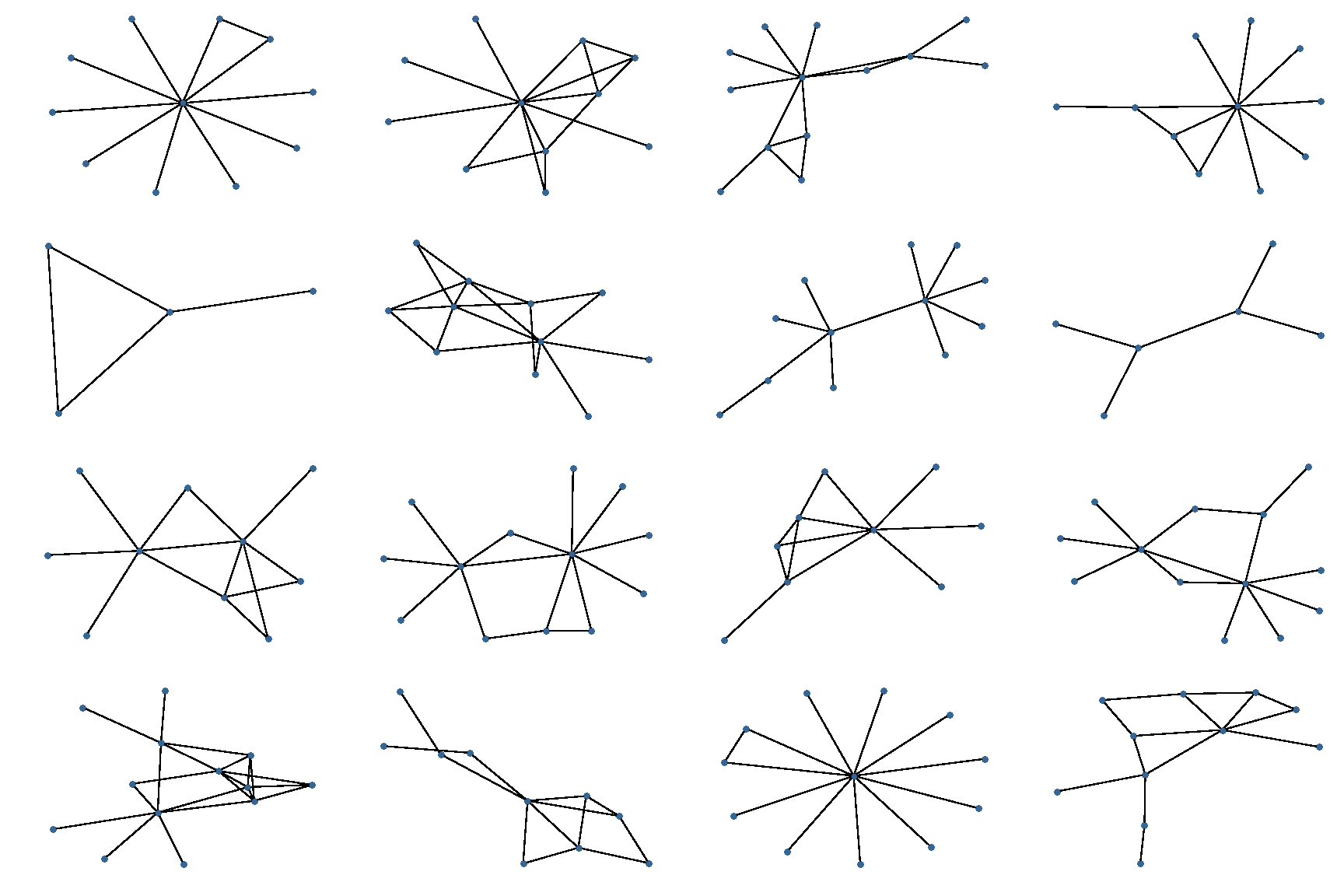} & \mfigure{width=0.25\linewidth}{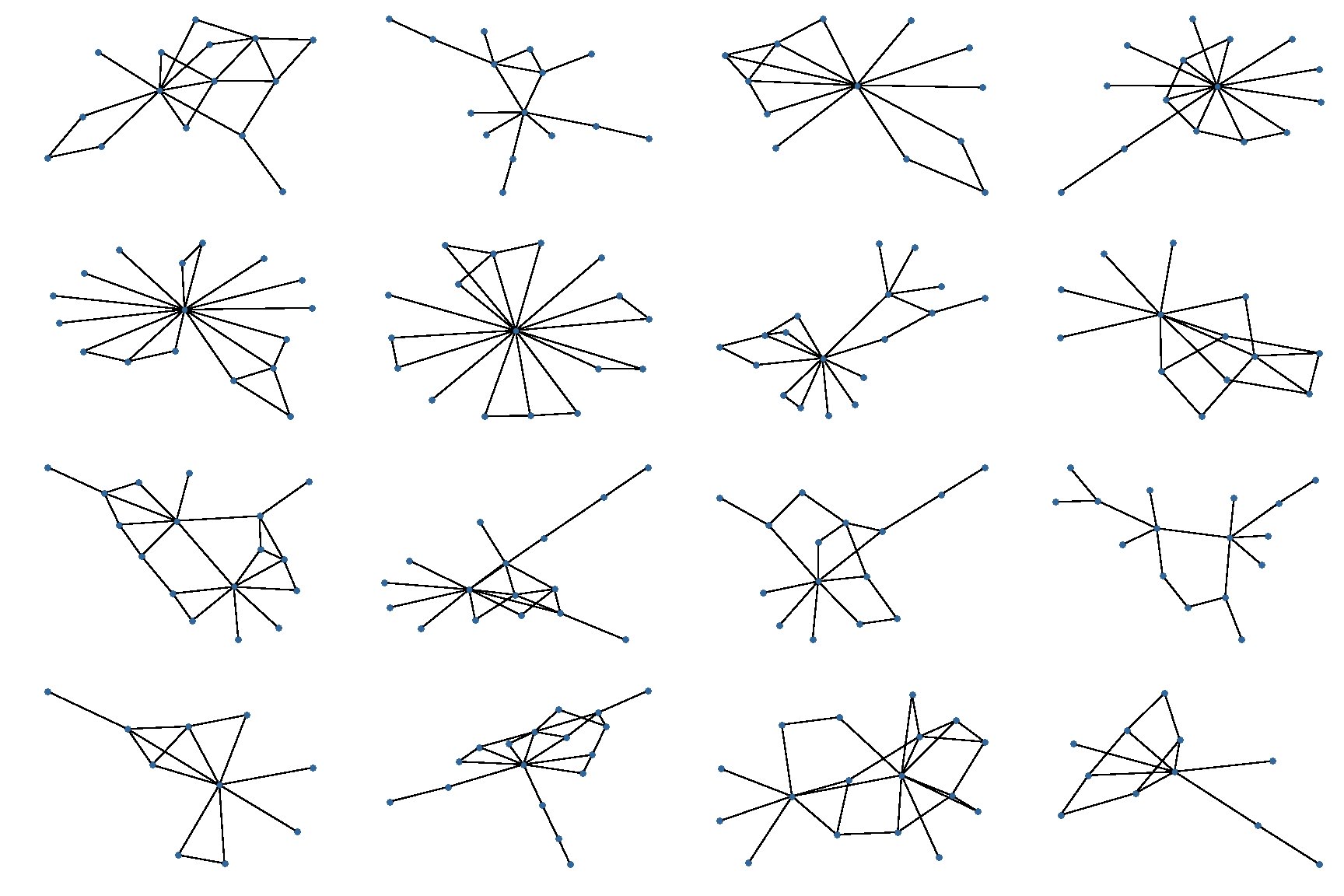} \\ 
         \midrule
         \multicolumn{3}{c}{Com-small} \\
         \mfigure{width=0.33\linewidth}{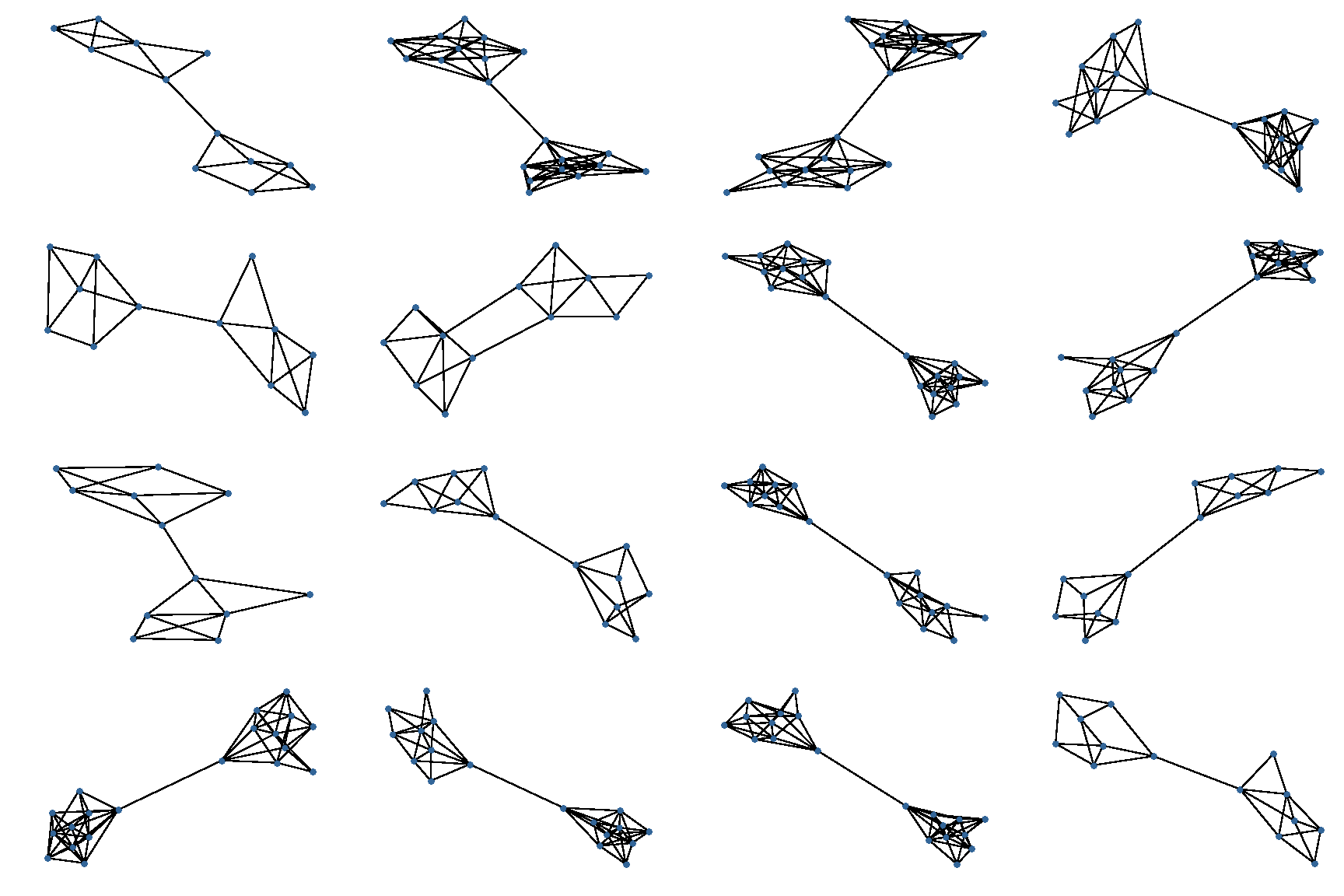} & \mfigure{width=0.33\linewidth}{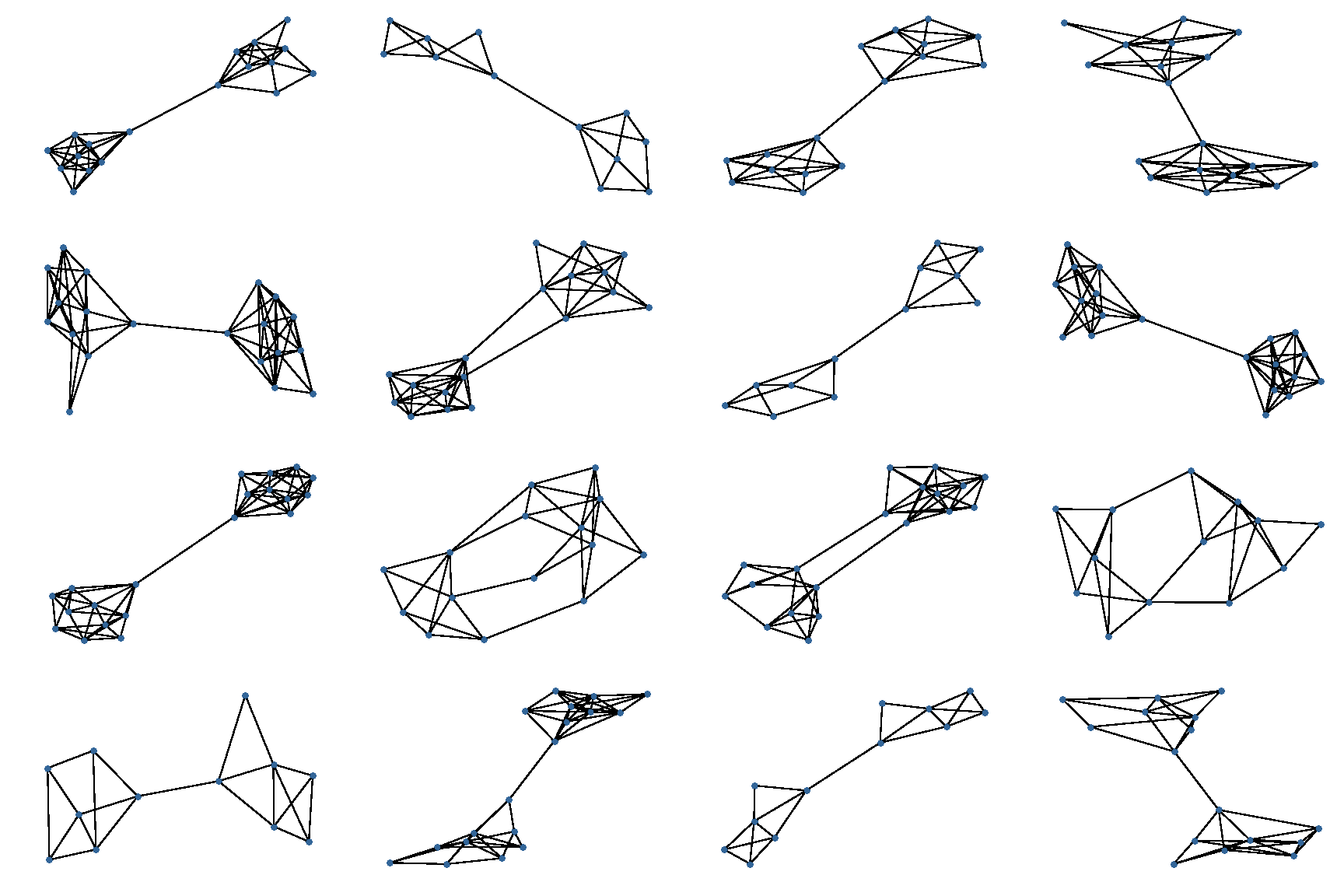} & \mfigure{width=0.33\linewidth}{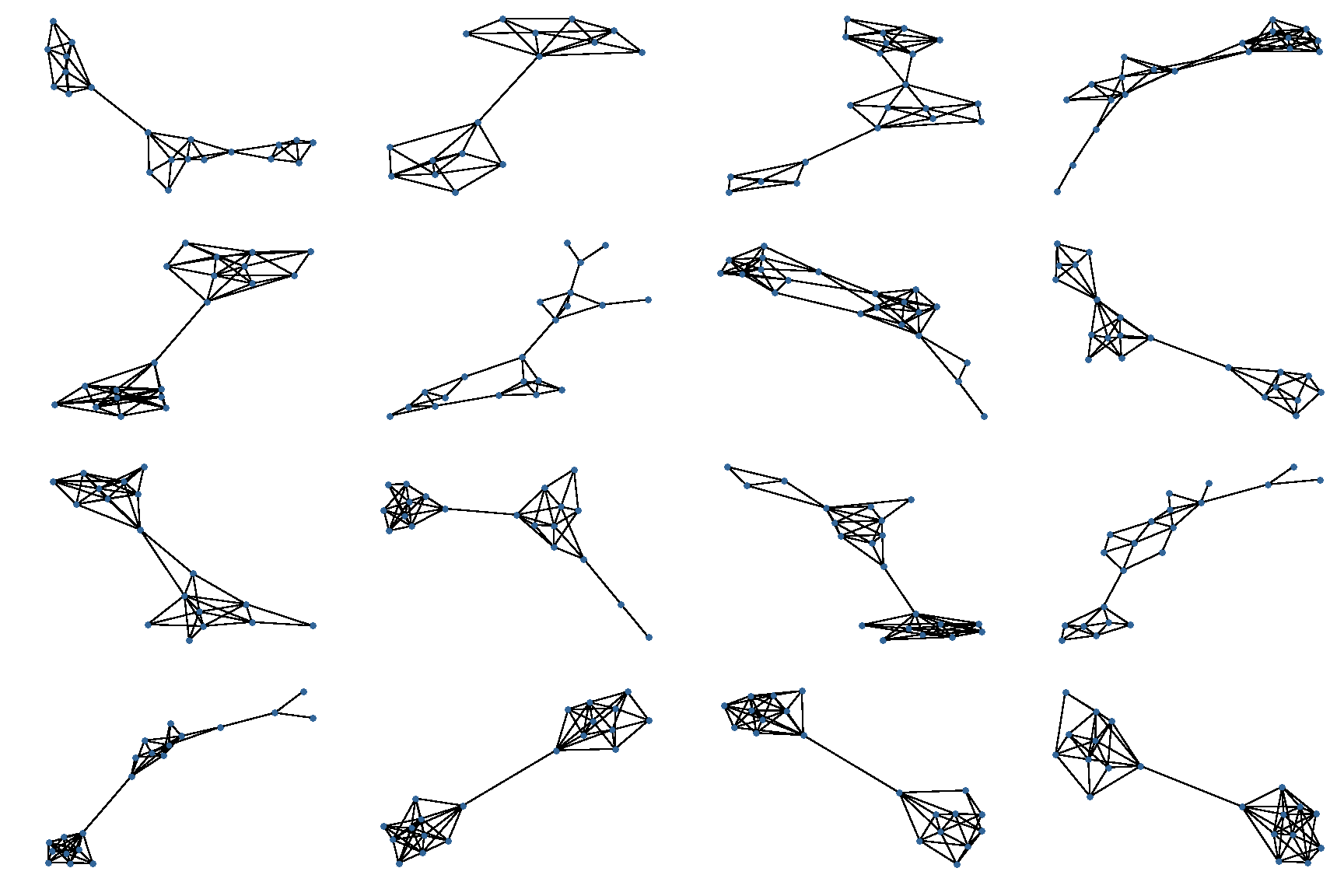} \\ 
        \midrule
        \multicolumn{3}{c}{Com-mix} \\
        \mfigure{width=0.33\linewidth}{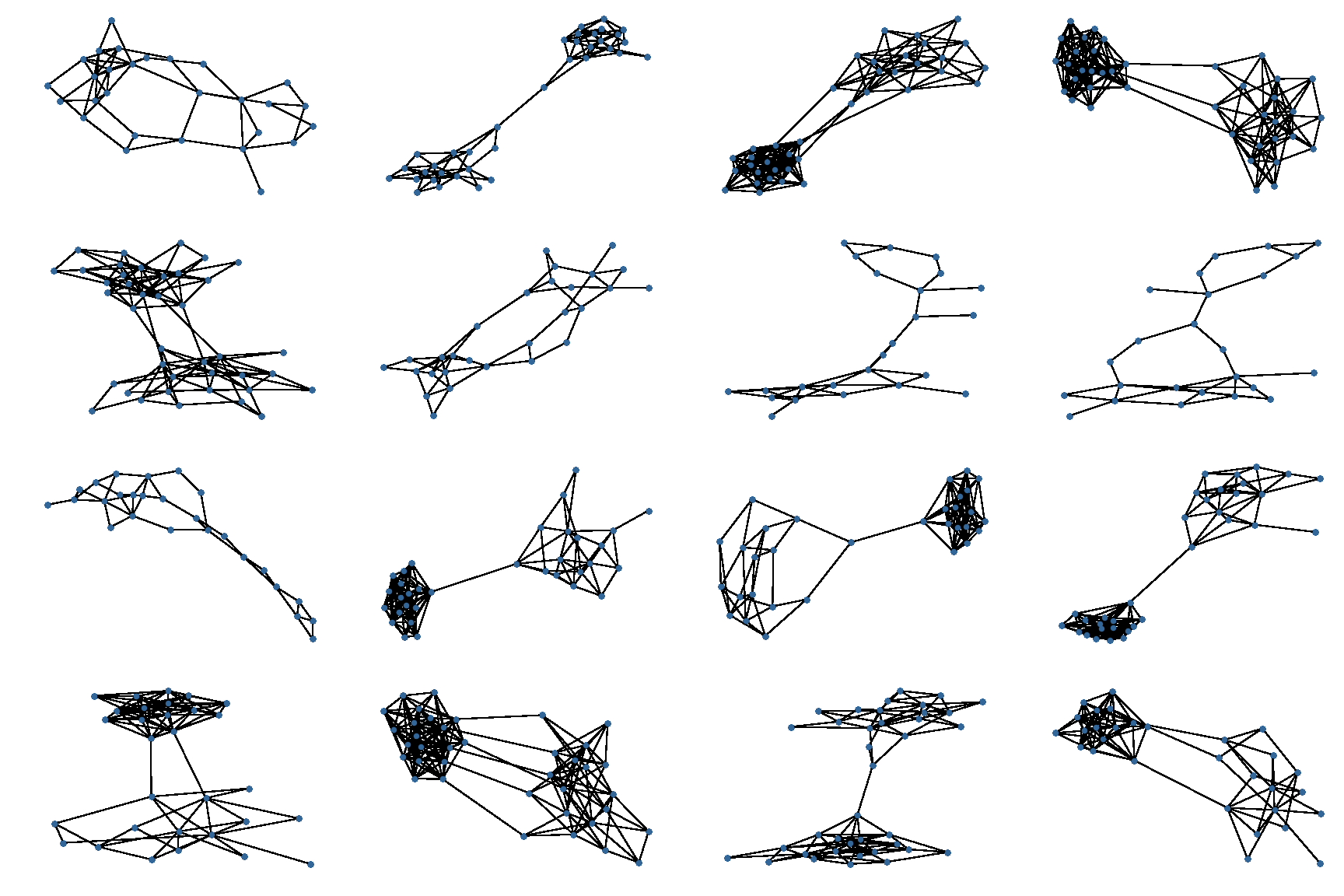} & \mfigure{width=0.33\linewidth}{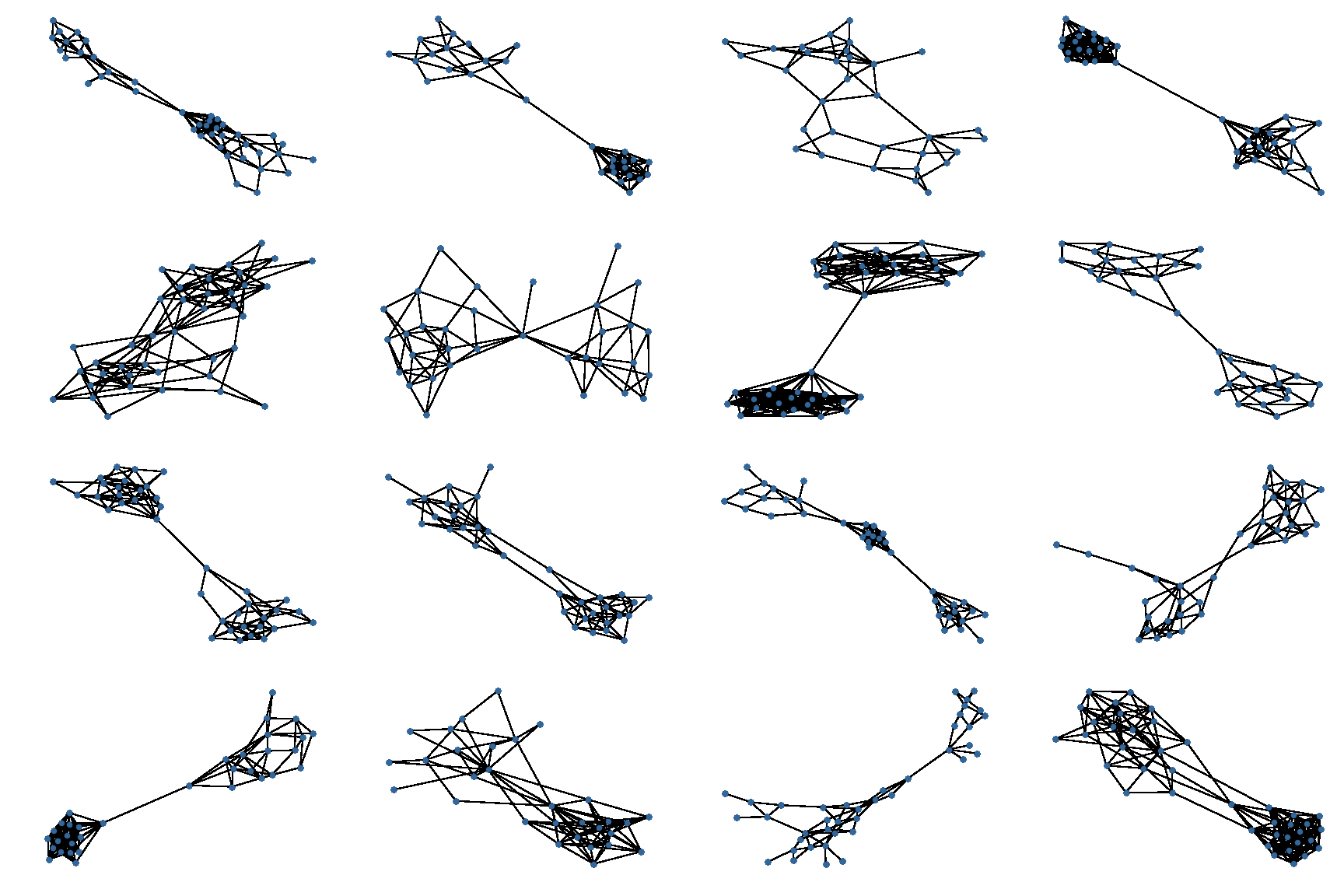} & \mfigure{width=0.33\linewidth}{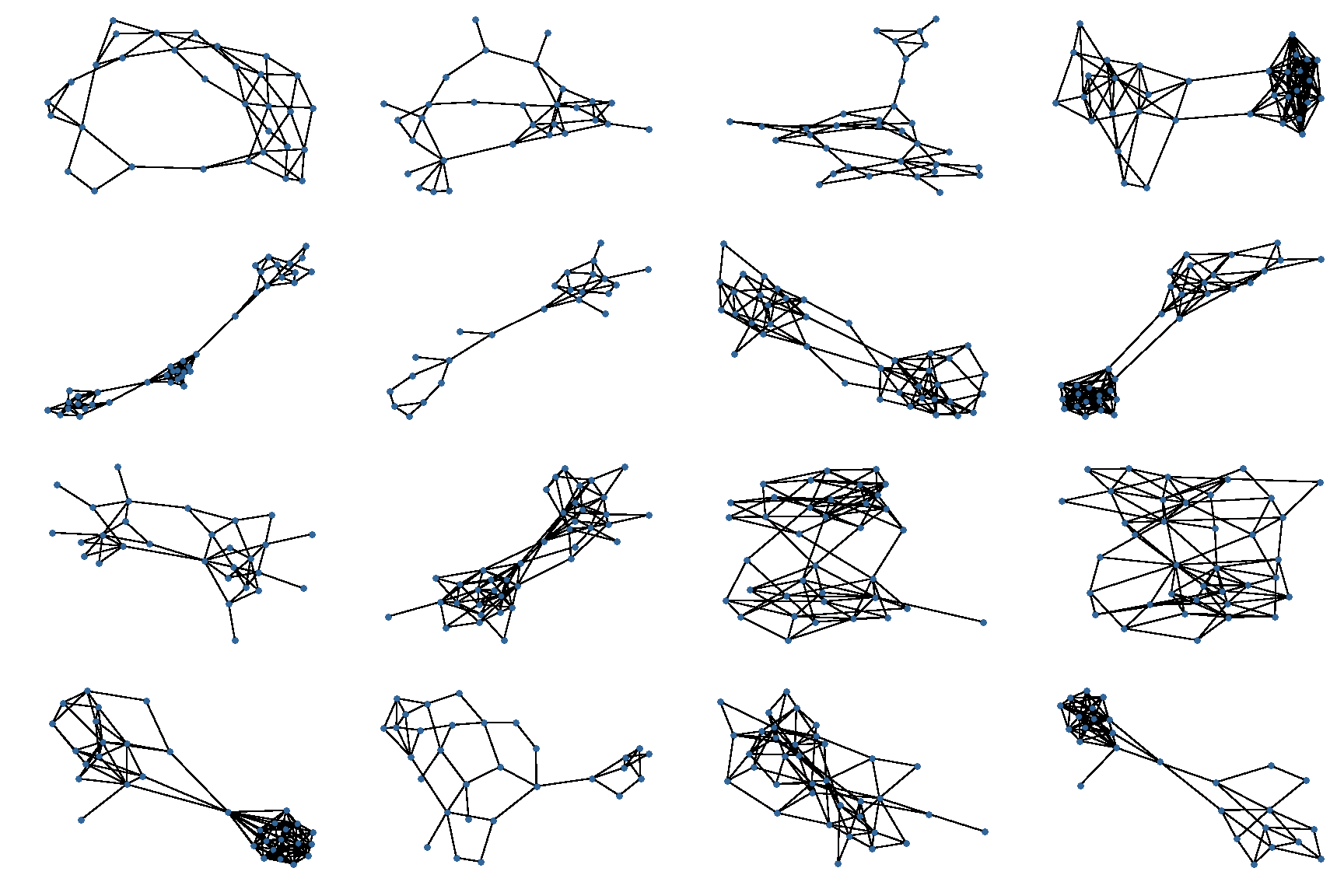} \\ 
        \midrule
        \multicolumn{3}{c}{Com-attr} \\
        \mfigure{width=0.33\linewidth}{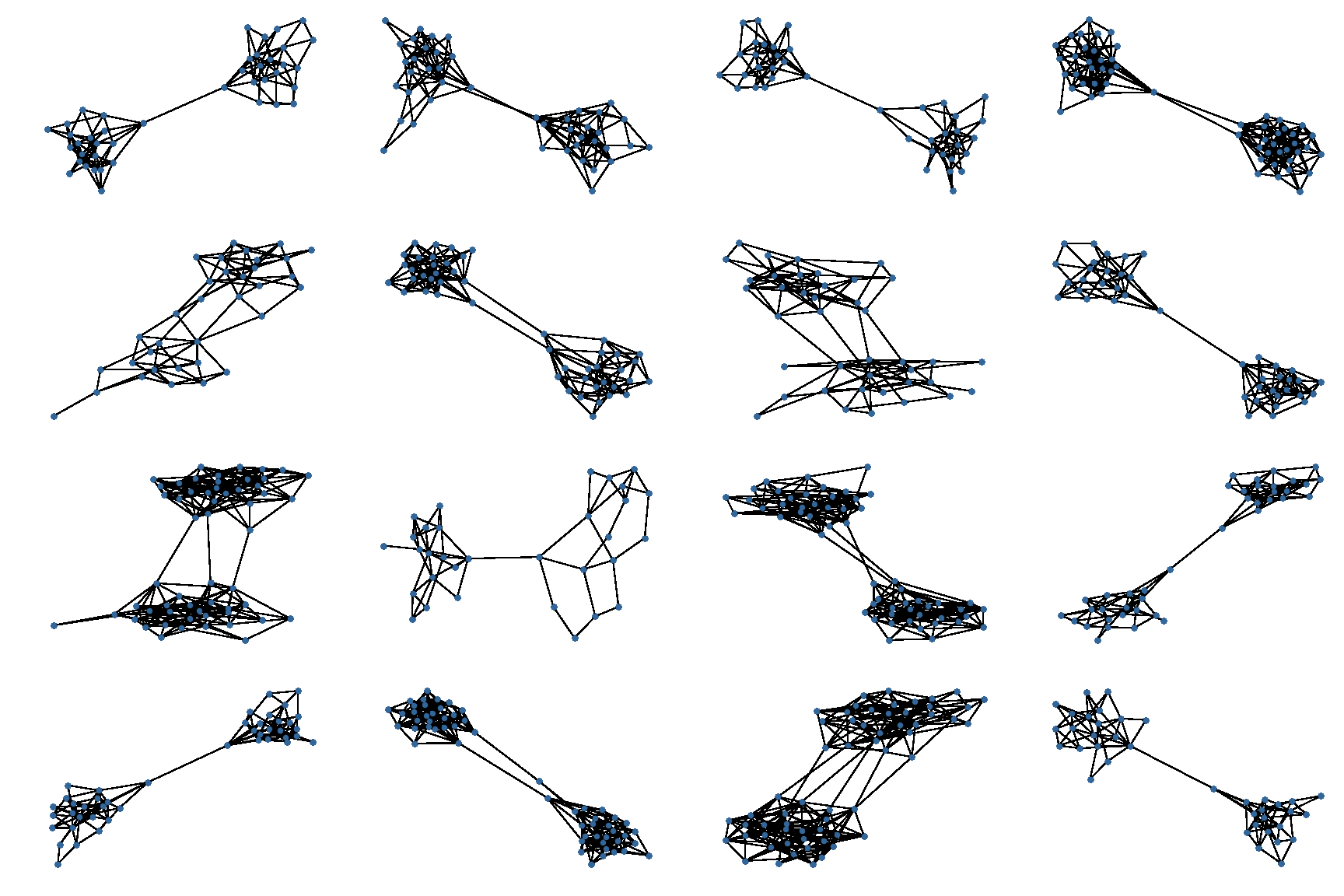} & \mfigure{width=0.33\linewidth}{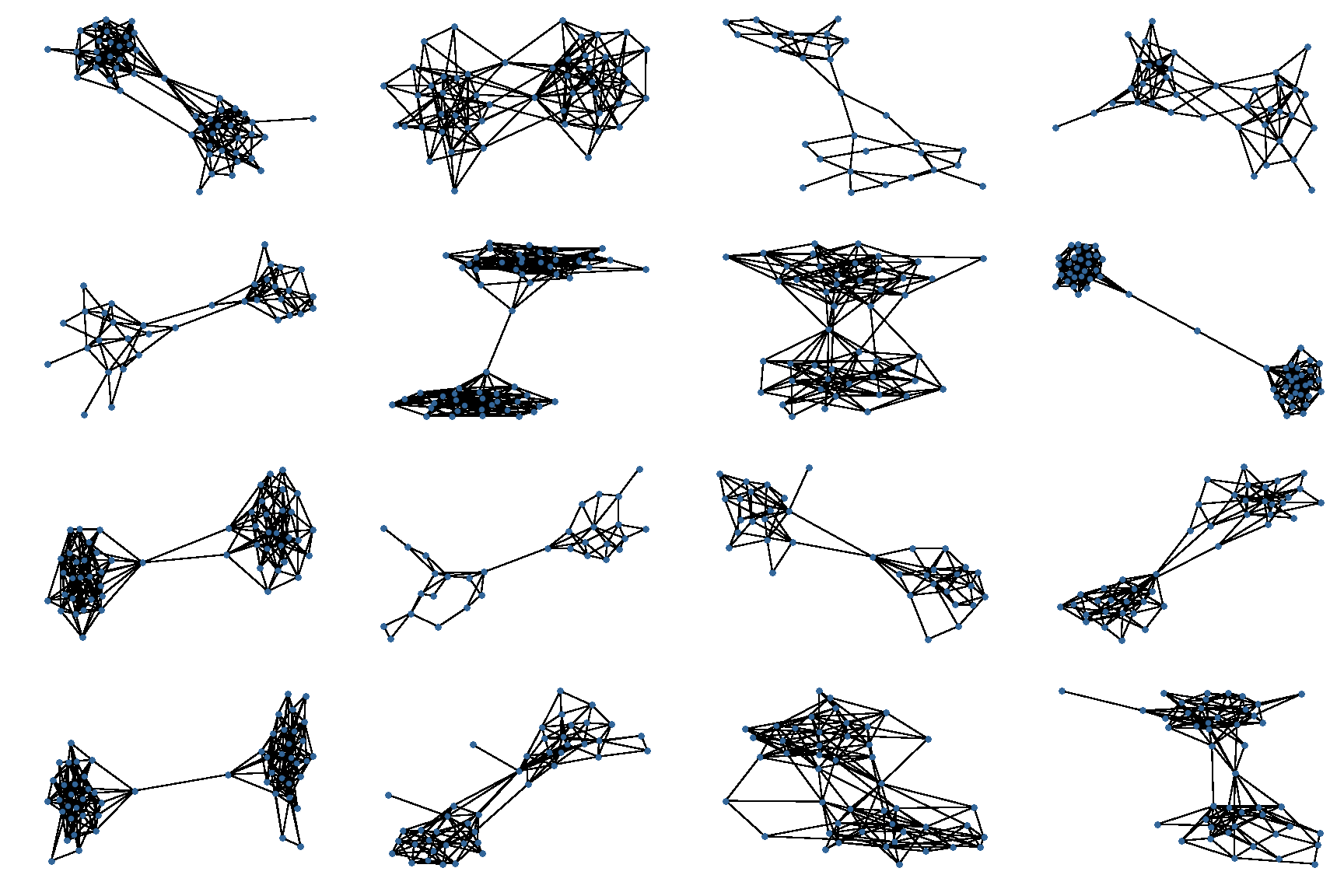} & \mfigure{width=0.33\linewidth}{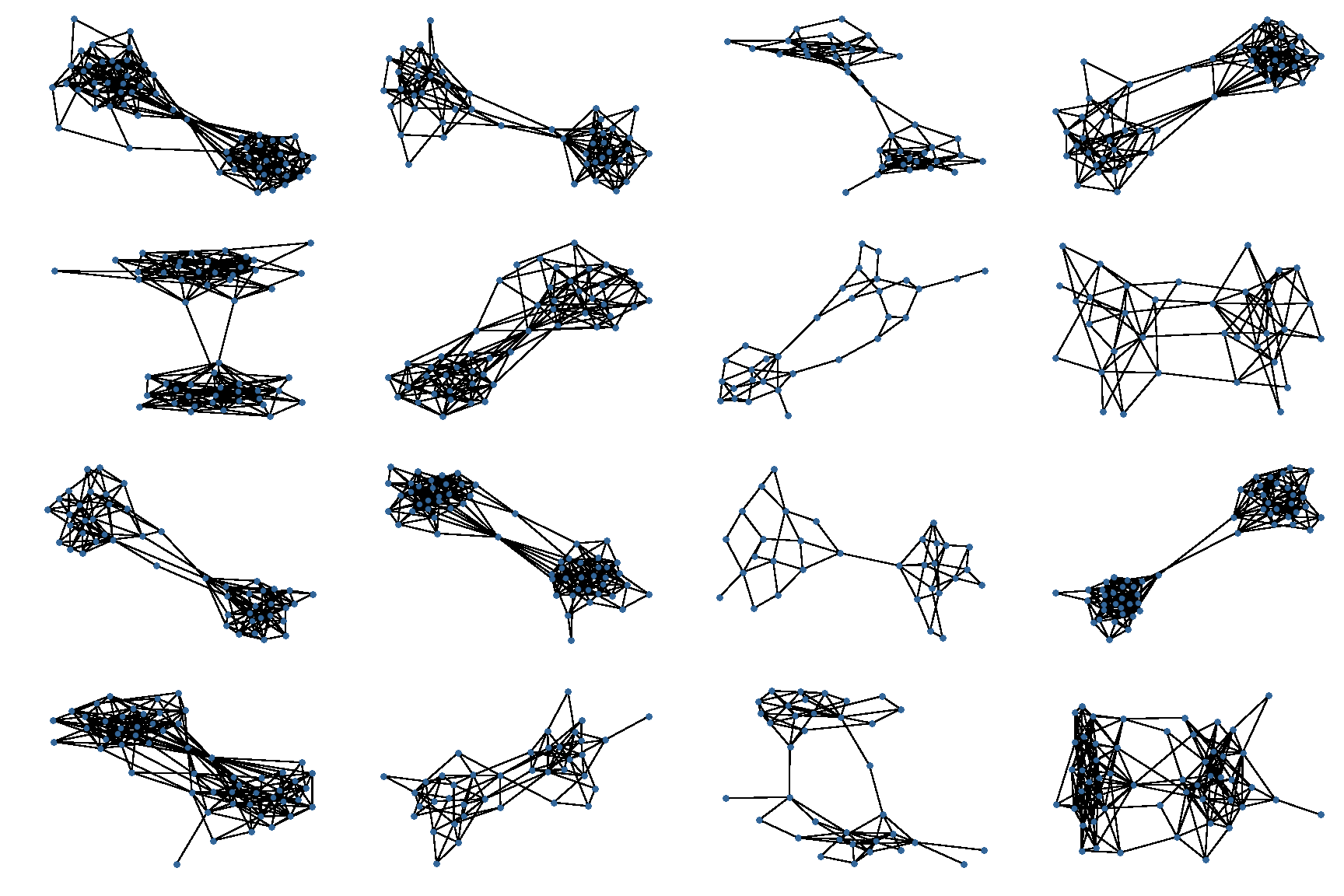} \\ 
         (a) Training samples& (b) GraphVRNN & (c) GraphRNN 
 \\ 
\end{tabular}
\caption{
\textbf{Qualitatitive results of the generated samples.} 
}
\label{fig:generated}
\end{figure}
\subsection{Dataset details}
In Figure~\ref{fig:dataset}, we show two illustrative examples of the 2-community datasets (namely Com-small, Com-mix and Com-attr) considered in Section~\ref{sec:exp}. For these datasets, each graph consists of two communities. The edge connection within each community is determined by the intra-connection probability, while the connection across the two communities is determined by the inter-connection probability. For Com-attr, the two communities have the same intra-connection probability 0.3 (and 0.7 for Com-small dataset), as shown in Figure~\ref{fig:comattr}. On the other hand, the intra-connection probability is not necessarily the same. As illustrated in Figure~\ref{fig:commix}, there exists two kinds of graphs in Com-mix: one with different intra-connection probabilities for the two communities (0.6 and 0.4), and the other with the same intra-connection probabilities. 

\subsection{Qualitative results}
We present the graphs generated from GraphVRNN and GraphRNN in Figure~\ref{fig:generated}. For each dataset, we pick the best model out of the 5 training runs, and randomly select 16 graphs from the generated samples.
\end{document}